\documentclass{article}
\usepackage{spconf,amsmath,graphicx}
\usepackage[export]{adjustbox}
\usepackage{url}
\usepackage{subfig}

\title{Embeddings for DNN speaker adaptive training}

\name{Joanna Rownicka, Peter Bell, Steve Renals\thanks{This work was supported by a PhD studentship from the DataLab Innovation Centre, Ericsson Media Services, and Quorate Technology.}}
\address{Centre for Speech Technology Research, University of Edinburgh, UK}

\begin{document}
\maketitle
\begin{abstract}

In this work, we investigate the use of embeddings for speaker-adaptive training of DNNs (DNN-SAT) focusing on a small amount of adaptation data per speaker.   DNN-SAT can be viewed as learning a mapping from each embedding to transformation parameters that are applied to the shared parameters of the DNN.  We investigate different approaches
to applying these transformations, and find that with a good training strategy, a multi-layer adaptation network applied to all hidden layers is no more effective than a single linear layer acting on the embeddings to transform the input features. In the second part of our work, we evaluate different embeddings (i-vectors, x-vectors and deep CNN embeddings) in an additional speaker recognition task in order to gain insight into what should characterize an embedding for DNN-SAT. We find the performance for speaker recognition of a given representation is not correlated with its ASR performance; in fact, ability to capture more speech attributes than just speaker identity was the most important characteristic of the embeddings for efficient DNN-SAT ASR. Our best models achieved relative WER gains of 4\% and 9\% over DNN baselines using speaker-level cepstral mean normalisation (CMN), and a fully speaker-independent model, respectively.

\end{abstract}
\section{Introduction}
\label{sec:intro}

The robustness of an automatic speech recognition (ASR) system can be enhanced by using various acoustic model adaptation methods, which aim to modify a general model towards particular testing conditions, or modify testing features towards a general model. However, as opposed to the adaptation of Gaussian mixture model -- hidden Markov model (GMM-HMM) systems, adaptation of deep neural networks (DNNs) still remains an open research question, due to lack of interpretability of the model parameters~\cite{book_new_era}.

In a speaker adaptively trained DNN (DNN-SAT), a speaker representation is usually extracted from a separate system and appended to the input acoustic features (e.g. mel frequency cepstral coefficients -- MFCCs), enabling the network to learn speaker-invariant representations~\cite{book_new_era,ivectors_2013} and improving the robustness.
Of course, embeddings may be extracted for any other labelled attribute of the data, and hence in this paper we refer more generally to attribute embeddings or attribute-aware training.
Attribute embeddings can be transformed with a control network prior to the concatenation with input features, to better learn attribute specific feature shifts~\cite{miao}.  Separate shift or scale transformation can also be learned from the embeddings for each hidden layer, and applied to internal feature representations at the output of any hidden layer of the main network to enhance the attribute invariance for the internal representations~\cite{swietojanski2016learning,subspace_LHUC,ibm_embed}. 

In this work, we explore how the transformation of the attribute embedding influences the attribute-normalization ability in the main part of the acoustic model, and whether the transformation to more abstract representations at the hidden layers is superior to the transformation of input features. We perform experiments with embeddings extracted at the utterance and speaker levels (extracted per frame), investigating the effectiveness of DNN-SAT with limited quantity of adaptation data.  To the best of our knowledge, online i-vectors have never been previously used for DNN-SAT to generate layer-wise transformation parameters.

A second contribution of this work is the analysis of the embeddings used for DNN-SAT. Building on deep speaker embedding extraction methods proposed for speaker recognition~\cite{xvectors,deep_speaker,deep_speaker_2017,deep_speaker_2018}, we used different speaker representations~\cite{online_ivectors,bsv,microsoft_cortana},  representations capturing different speech attributes (e.g. noise)~\cite{ivectors_really,noise_vector}, and purely utterance-level summaries~\cite{Brno,dynamic_layer,dialect} for DNN-SAT. The premise is that with better attribute representations, itis possible to learn more reliable transformations, enabling the network to operate in a feature space better normalized with respect to the attribute. We investigate the characteristics of three types of embeddings (i-vectors, x-vectors, and deep  convolutional neural network  (CNN) embeddings~\cite{my_slt}) by evaluating them for speaker recognition task, and then using the same embeddings DNN-SAT ASR.  To our knowledge, x-vectors have not been previously explored for SAT.  These experiments enable us to relate speaker discriminability of each embedding to the performance in DNN-SAT, in order to better understand which properties are desirable for efficient adaptation, serving as guidance for designing new embeddings for this task.

\section{Embeddings incorporation into acoustic model}

We explore different ways of incorporating the embeddings into the acoustic model to enhance its attribute invariance. We start from a multi-layer control network (sec.~\ref{sec:control_net}) used to map each embedding into shift and scale parameters to (a) normalize all hidden representations, and (b) to normalize input features. We then reduce the number of parameters acting on the embeddings with a control layer (sec.~\ref{sec:embed_layer}), control vector (sec.~\ref{sec:embed_vec}) and control variable and constant scale (sec.~\ref{sec:embed_scale}). Element-wise addition is denoted with $\bigoplus$ (shifting), and $\bigodot$~denotes element-wise product (scaling).

\begin{figure}[t]
    \includegraphics[scale=0.45,center]{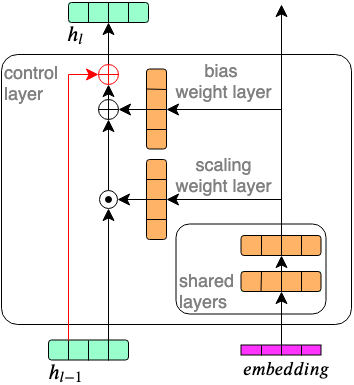}
    \caption{Embedding-based DNN-SAT using control network with a skip connection (red) to learn the shift and scale to all hidden layers of the main part of the acoustic model.}
    \label{fig:control_network}
\end{figure}

\subsection{Control network}
\label{sec:control_net}

Several approaches have been proposed recently to incorporate an attribute embedding into the network to normalize the hidden representations. In~\cite{ibm_embed} i-vectors are mapped through a network to element-wise scaling and
bias parameters. Our first approach to incorporate the embedding into acoustic model is similar to~\cite{ibm_embed} and is depicted in Figure~\ref{fig:control_network}. In our work, however, attribute-dependent mappings are generated by a control network from online i-vectors, not speaker i-vectors, to enable efficient DNN-SAT. During decoding, the DNN-SAT model is adapted simply by
extracting online i-vectors of each test speaker at the frame level, feeding the i-vector forward through the adaptation network, and integrating the
generated shift and scale with the internal representations of the main part of the network. Differently to~\cite{ibm_embed}, we also add a skip connection to the control network, in order to control the degree of adaptation at each layer. By doing this, instead of applying a control network just to one or two bottom layers of the main network, we may insert the control network after every layer without a loss in performance. Shared layers of the control network have 100 units with ReLU nonlinearity and are applied to an externally extracted speaker embedding (online i-vector). Scaling and bias weight layers are used to scale and shift hidden representations ($\mathbf{h_{l-1}}$), and are thus constrained to have the same number of units as the layers in the main part of the network.  The scaling weight layer uses a sigmoid activation function to enforce positive scaling whilst the bias weight layer employs a hyperbolic tangent activation, enabling the bias to be positive or negative. Those layers act on the output of the shared layers and are applied to hidden representation $\mathbf{h_{l-1}}$ to obtain normalized hidden representation $\mathbf{h_l}$. We apply the control network at each layer $l$ of the main network to learn the shift and scale to all of the hidden layers. We also experiment with applying the control network only to the input features.

\subsection{Control layer}
\label{sec:embed_layer}

\begin{figure*}[thbp]
\centering
\null\hfill
\subfloat[Control layer]{\includegraphics[scale=0.45]{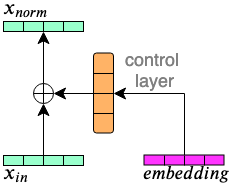}}
\hfill
\subfloat[Control vector]{\includegraphics[scale=0.45]{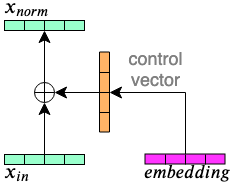}}
\hfill
\subfloat[Control variable or constant scale]{\includegraphics[scale=0.45]{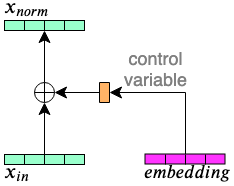}}
\hfill\null
\caption{Approaches to incorporate the embeddings into the acoustic model, reducing the number of parameters acting on the embeddings compared to the control network.}
\label{fig:layer_vec_var}
\end{figure*}

Instead of using a multi-layer control network, we reduce the number of parameters acting on the embeddings by using a single control layer to transform the representations (Figure~\ref{fig:layer_vec_var}a), to avoid overfitting. With a control layer, we apply the shift or scale to the input features, $\mathbf{x_{in}}$, or all hidden representations. The equations and figures in sections~\ref{sec:embed_layer}-\ref{sec:embed_scale} show input features transformations. For hidden representations transformation, $\mathbf{x_{in}}$ is replaced with $\mathbf{h_{l-1}}$ and $\mathbf{x_{norm}}$ with $h_{l}$. Feature shifts (Eq.~\ref{eq:layer}) or scalings are generated from the embeddings, $\mathbf{e}$, to obtain normalized features, $\mathbf{x_{norm}}$, that are then used in the normalized feature space by the main part of the network.

\begin{equation}
  \mathbf{x_{norm}} =  \mathbf{x_{in}} + f(\mathbf{e})
  \label{eq:layer}
\end{equation}

\begin{equation}
   f(\mathbf{e}) = act( \mathbf{W_e}^T \mathbf{e} + \mathbf{b_e} )
  \label{eq:layer2}
\end{equation}

In our experiments, we use the ReLU, sigmoid, tanh and linear activation functions (\emph{act}) to explore the effect of the linear vs. nonlinear shift, and the direction and value range of the shifts to the input features. $\mathbf{W_e}$ and $\mathbf{b_e}$ are the weight matrix and bias vector acting on the embeddings. Feature scaling can be realized by replacing addition with element-wise product in Eq.~\ref{eq:layer} and in Figure~\ref{fig:layer_vec_var}a. 

\subsection{Control vector}
\label{sec:embed_vec}

To further reduce the number of parameters applied to the embeddings, we estimate just the diagonal elements of the weight matrix $\mathbf{W_e}$:

\begin{equation}
  \mathbf{W_e} = diag(\mathbf{w_e})
  \label{eq:vector1}
\end{equation}

The resulting vector $\mathbf{w_e}$ can then be used to scale the embedding $\mathbf{e}$, and shift the input features (Figure~\ref{fig:layer_vec_var}b). Here, we use the sigmoid activation function to restrict the scaling of the embeddings to only positive values.

\begin{equation}
  \mathbf{x_{norm}} =  \mathbf{x_{in}} + sigmoid(\mathbf{w_e}) \otimes \mathbf{e}
  \label{eq:vector2}
\end{equation}

\subsection{Control variable or constant scale}
\label{sec:embed_scale}

To further simplify the approach and reduce the number of parameters, a single weight $w_e$ can be used to scale the embedding prior to shifting the input features (Figure~\ref{fig:layer_vec_var}c). 

\begin{equation}
  \mathbf{x_{norm}} =  \mathbf{x_{in}} + w_e \cdot \mathbf{e}
  \label{eq:variable}
\end{equation}

Moreover, to eliminate the need to learn any parameters acting on the embeddings, we also keep the scale constant (0.1) instead of learning the weight $w_e$.
Finally, the concatenation of the input features and the embeddings can be seen as weighting the embedding with a constant scale equal to 1. In section~\ref{sec:incorporation} we show empirical results for all mentioned approaches to map the online speaker i-vectors to control parameters used to transform input features as well as hidden representations.

\section{Results and discussions}
\label{sec:results}

\subsection{Experimental setup}
\label{sec:setup}

In all our experiments we use the AMI IHM dataset~\cite{ami} which contains around 100 hours of meeting recordings in English. 
We use the Kaldi toolkit~\cite{kaldi} for input acoustic feature extraction (40-dim high-resolution MFCC features with double deltas and 5 frames of context at each side, with cepstral mean normalisation (CMN) applied except where specified), i-vector and x-vector extraction, training initial models for alignment, and decoding. 
We use the PyTorch-Kaldi~\cite{pytorch-kaldi} toolkit to implement DNN acoustic models and Tensorflow~\cite{tensorflow} to extract deep CNN embeddings.

For our ASR experiments, we use the train/dev/eval set split defined by Kaldi recipe\footnote{\url{egs/ami/s5b/local/chain/run_tdnn.sh}}.
Unlike the recipe we do not employ any speed or volume perturbation to the training set. The main acoustic model is a 6 layer DNN with 2048 units in each layer, trained with cross-entropy loss over 3984 context-dependent tied triphone states. 
For decoding we use the trigram language model from the standard recipe, which is an interpolation of the trigram language models trained on the AMI and Fisher English transcripts.   
  
To explore the characteristics of the embeddings, we evaluate them in a speaker recognition task. i-vectors used in this work are extracted from MFCCs in an online fashion or at the utterance level. Online i-vectors are extracted at each frame; however, the speaker information is carried over within speakers. We explore both types of i-vectors for efficient DNN-SAT. i-vectors are designed to capture both speaker and channel characteristics, as they use a single variability subspace to model different types of variability in the speech signal. x-vectors are extracted per utterance; however, since the x-vector extractor is trained with the use of speaker labels, they are explicitly designed to capture speaker characteristics. Compared to i-vectors, x-vectors should therefore be invariant to within-speaker channel variability. Deep CNN embeddings are also extracted per utterance. Here, the speaker labels are not used in the embedding extraction. The model used to extract the embeddings is a very deep CNN acoustic model~\cite{my_simply} (similar to the VGG~\cite{vgg} architecture but without pooling layers) with 2D 3x3 kernels, trained to classify senone states. 

Principal components analysis (PCA) is used to reduce the dimensionality of the embeddings. To add speaker discrimination to the embeddings, a linear discriminant analysis (LDA) transform informed by training speaker labels may be used. It was shown in~\cite{my_slt} that PCA CNN embeddings are much more characteristic of the acoustic condition than i-vectors (for Aurora4 dataset), and LDA CNN embeddings are also better speaker representations, compared to i-vectors.
In the speaker recognition experiments, we create our own split of data for enrollment (enroll) and testing (test). We first merge the original dev and eval sets (Kaldi ASR split) to maximize the number of speakers for evaluation (135), and then split the set into two parts, such that utterances from every speaker are found in both enroll and test sets. We use the enroll set to obtain speaker-level representations, and we evaluate the utterance-level representations from the test set against them. We create the trials with non-target proportion 50\% and with the non-matching speaker for the non-target part of trials chosen at random. 

\subsection{Embedding incorporation}
\label{sec:incorporation}

\begin{table}[t]
    \centering
    \begin{tabular}{l|l|r}
       \hline
       \emph{Embed. mapping} & shift/scale & WER \\
       \hline
       SI baseline & - & 28.3 \\ 
       CMN & - & 27.0 \\  
       \hline
       ctrl network & $\bigoplus$, $\bigodot$ & 27.0 \\
       ctrl layer & $\bigodot$ & 26.5 \\ 
       ctrl layer & $\bigoplus$ & \textbf{25.9} \\ 
       ctrl vector & $\bigoplus$ & 26.1 \\ 
       ctrl variable & $\bigoplus$ & 26.1 \\ 
       ctrl scale & $\bigoplus$ & 26.1 \\ 
       embed. concat. & & 26.5 \\ 
       \hline
       ctrl network (hidden) & $\bigoplus$, $\bigodot$ & 26.1 \\  
       ctrl network (hidden) & $\bigoplus$ & 26.0 \\ 
       ctrl network (hidden) & $\bigodot$ & 26.0 \\  
       ctrl layer (hidden) & $\bigoplus$ & 26.4 \\ 
       ctrl layer (hidden) & $\bigodot$ & 27.4 \\ 
       \hline
    \end{tabular}
    \caption{Comparison of the approaches to generate the parameters acting on the embeddings for input features and hidden representations transformation. All models (except for the SI baseline) use CMN features.}
    \label{tab:incorp}
\end{table}

The results for different embedding mapping and transformation approaches are presented in Table~\ref{tab:incorp}. The first block gives baseline results, the second one is for input features transformation, and the third is for hidden representation transformations. Different patterns can be observed depending on where in the network the transformation is applied. For input feature transformation, a multi-layer control network did not outperform the CMN baseline. All of the approaches with fewer parameters were superior to the control network.  Since the control network is the most flexible, this suggests that when the control network is applied as low in the network as to the input layer the model may be more vulnerable to overfitting.

For the control layer, shifting the input features was more important than scaling them with the activations generated from the embeddings. The input features are already mean-normalized per speaker, so the transformations are potentially normalizing different factors of variation in the utterances. This could explain better performance of shifting rather than scaling with the control layer -- additive noise might be compensated by shifting. 
For the control layer acting on the embeddings, we experimented with different activation functions, finding that all of them give similar performance. The simplest and at the same time the most flexible approach is to use a linear activation function. It does not restrict the direction and the value range of the feature shifts. When using a linear identity activation function we do not benefit from learning more abstract embedding representations prior to the input feature transformation -- so the advantage of this approach lies in scaling the embeddings with the weights learned in the control layer. 

Furthermore, we found  the training strategy to be important. The most effective approach was to initialize the main part of the network with the parameters from the SI baseline in the first stage, and in the second stage to train the remaining control parameters, together with fine-tuning the parameters of the main part of the network. This approach was much better than fixing the parameters of the main part of the network in the final stage and only training the control part of the network. We hypothesize that this is due to the ability to update the parameters of the DNN model in the new, normalized feature space. The training strategy is the factor differentiating the embedding concatenation vs. other approaches. It is therefore important to scale the embedding, as well as to use a multi-step training approach to generate the control parameters for the input feature transformation in the attribute-aware DNN-SAT scheme.

The results for the transformations applied to the hidden layers (third block) show that, overall, transforming hidden representations is not more effective than transforming the input features. This might be explained by the fact that batch normalization is used in our experiments, so the hidden representations might be already invariant to different speech attributes. Further transformation with control parameters derived from the embeddings is not as effective as transforming the input features, allowing the main part of the network to operate in the normalized feature space at the first hidden layer. Interestingly, the hidden layers benefit from a different mapping approach to the input layers: in this case, the control network with the most flexibility to generate transformation parameters is superior to other approaches.  We conclude that learning an abstract layer-wise mapping is important for hidden layer transformations.

\subsection{Characteristics of the embeddings}
\label{sec:characteristics}

To better understand the characteristics of the embeddings and how they relate to ASR performance, we first evaluate them for speaker recognition, and then in the DNN-SAT setup with the control layer to shift the input features. The aim is to learn if better speaker representations extracted from NNs correlate with better performance in DNN-SAT.

\begin{table}[t]
        \caption{\label{tab:ami_spk} {EERs (\%) for i-vectors (per utt.), x-vectors, and deep embeddings evaluated in speaker recognition task with different backends. \emph{x-vector} uses a model trained on AMI training data. \emph{x-vector (pretrained)} uses Kaldi pretrained SRE16 x-vector model. \emph{x-vector (AMI backend)} also uses the pretrained model but the LDA and PLDA transforms are trained on the AMI training data.}}
        \vspace{2mm}
        \centerline{
          \begin{tabular}{l|cccc}
            \hline
              & cosine & PLDA & LDA & LDA/ \\
              & & & & PLDA \\
            \hline
            \textsl{400D i-vector} & \textbf{10.05} & 10.93 & 10.98 & 11.88 \\
            \textsl{x-vector} & 31.47 & 44.95 & 43.27 & 27.85 \\ 
            \textsl{x-vector (pretrained)} & - & - & 43.47 & 33.83 \\  
            \textsl{x-vector (AMI backend)} & 37.36 & 13.42 & 14.65 & 14.51 \\ 
            \textsl{400D CNN embed.} & 20.65 & 5.10 & 7.09 & 5.55 \\ 
            \textsl{800D CNN embed.} & 20.62 & \textbf{5.06} & \textbf{6.70} & \textbf{5.34} \\ 
			\hline
          \end{tabular}
       }
\end{table}

\begin{figure}[t]
    \centering
    \includegraphics[scale=0.48]{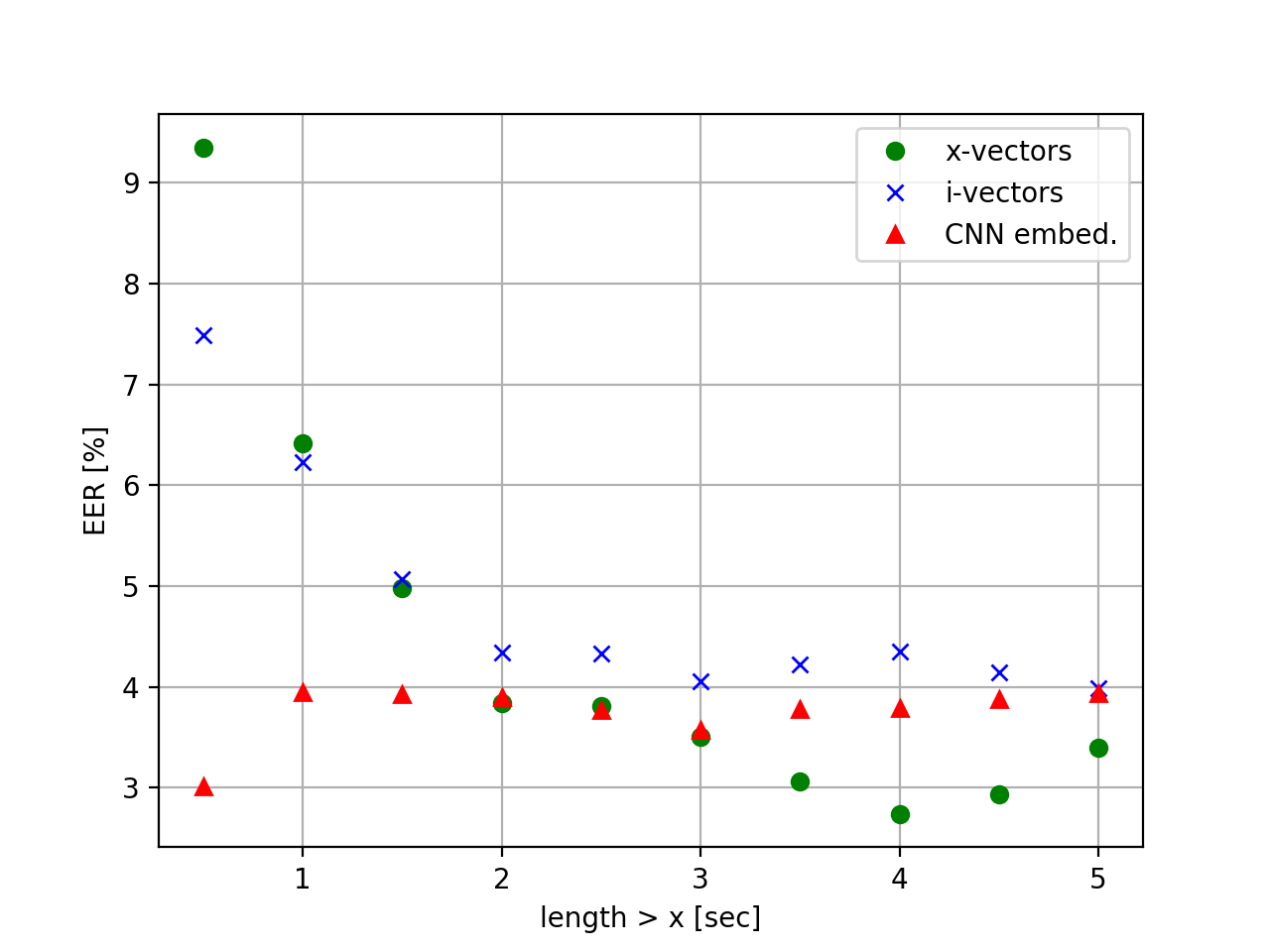}
    \caption{EERs (\%) for x-vectors, i-vectors and CNN embeddings with PLDA scoring for speaker recognition. Enroll and test recordings are filtered by the minimum length.}
    \label{fig:plda_xvector}
\end{figure}

The results for speaker recognition are presented in Table~\ref{tab:ami_spk}. We use the Equal Error Rate (EER) metric.  Deep CNN embeddings extracted with a 800D PCA transformation are the most accurate speaker representations. Comparing i-vectors and x-vectors, we find that i-vectors are superior with all backends. Even when extracted from a pretrained SRE16 x-vector model~\cite{xvectors} with LDA and probabilistic LDA (PLDA) transformations trained on matched AMI training data, denoted \emph{x-vector (AMI backend)}, x-vectors are still worse for speaker differentiation on the AMI dataset. However, it can be seen in Figure~\ref{fig:plda_xvector}, that for longer recordings, x-vectors start to outperform both i-vectors and deep CNN embeddings in the speaker recognition task. It is interesting to note that the trend in Figure~\ref{fig:plda_xvector} for deep CNN embeddings is different to that for i-vectors and x-vectors. For very short utterances ($<$ 1 sec), deep CNN embeddings are able to capture speaker identity better than for longer utterances.

This behaviour might be explained by the fact that the extraction of deep CNN utterance-level embeddings from a fully-convolutional architecture, with kernels spanning over time and frequency, enables the capture not only of speaker characteristics, but also of more local time and frequency acoustic patterns. This hypothesis is supported by the results in Table~\ref{tab:ami_spk_max30s}, where instead of speaker classification, we perform speaker subset classification. Speaker subsets are created by splitting within existing speakers into several groups, with a 30 sec of data per subset restriction. Contiguous utterances are assigned to the subsets. As in the previous experiment, the opposite trend can be observed for deep CNN embeddings than for the i-vectors -- EERs are lower for CNN embeddings and higher for i-vectors for speaker subsets, compared to genuine speakers. Therefore, deep CNN embeddings are not only very good at capturing speaker characteristics, but can also represent more local characteristics of the utterances, potentially corresponding to different acoustic conditions, channel characteristics, or phonetic content of the utterances. In the next section, we analyze how those characteristics correspond to the performance in DNN-SAT.

\begin{table}[t]
        \caption{\label{tab:ami_spk_max30s} {EERs (\%) for i-vectors and deep embeddings evaluated in the ``speaker'' recognition task with different backends, where speakers are split such that there is no more than 30 sec of data per speaker.}}
        \vspace{2mm}
        \centerline{
          \begin{tabular}{l|cccc}
            \hline
              & cosine & PLDA & LDA & LDA/ \\
              & & & & PLDA \\
            \hline
            \textsl{i-vector (MFCC)} & \textbf{14.99} & 13.13 & 12.83 & 13.54 \\ 
            \textsl{800D CNN embed.} & 31.32 & \textbf{3.46} & \textbf{4.36} & \textbf{3.19}  \\
			\hline
          \end{tabular}
       }
\end{table}

\subsection{Embeddings for DNN-SAT}
\label{sec:characteristics_sat}


We evaluate the influence of the type of embedding and its characteristics on the DNN-SAT task, using a control layer to shift the input features. The results are presented in Table~\ref{tab:different_embeddings}. By replacing online i-vectors with offline utterance i-vectors, we gained 0.1\% in WER, thus the frame level variation of the embeddings did not contribute to better WER in our experiments. On the other hand, online decoding can only be performed with online representations, and as the loss of performance is not substantial, one may prefer to use online i-vectors. It is important to bear in mind however, that genuine speaker labels were used for online i-vector extraction in our experiments. For the deployment of DNN-SAT with online representations, an additional diarization or speaker clustering step would be needed. 

The PCA transform in the deep CNN embedding extraction framework is used simply to reduce the dimensionality of the embedding in an unsupervised fashion. Since speaker information is not used in the embedding extraction, PCA CNN embeddings can be regarded as utterance summaries, capturing local acoustic characteristics. Their incorporation into the acoustic model gives a substantial improvement over the baselines; however, adding speaker discrimination to the utterance summary (using LDA) further improves the results.
Interestingly, better speaker discriminability does not guarantee better normalization in DNN-SAT. The lowest EER was achieved with 20D embeddings for speaker recognition, and yet 100D embeddings were optimal for DNN-SAT. It is possible that by using more dimensions for the embeddings, we are able to capture more attributes of the utterance, resulting in better DNN-SAT performance. 

x-vectors are extracted from a network trained to classify speakers, hence they are more invariant to within-speaker variability, caused by speech attributes other than speaker identity. We believe that this might the reason for poorer ASR performance with DNN-SAT.

\begin{figure}[t]
    \centering
    \includegraphics[scale=0.51]{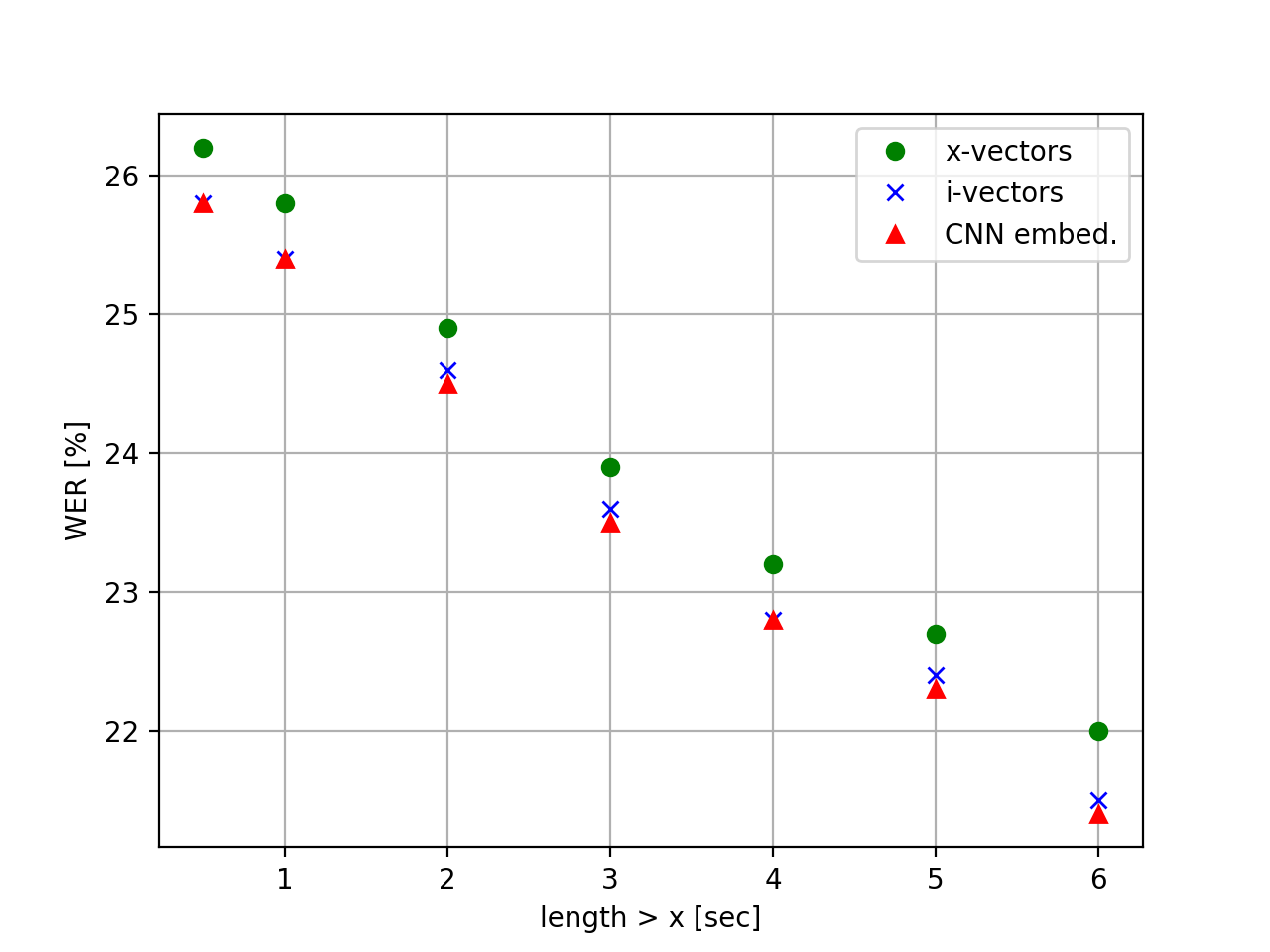}
    \caption{WERs (eval; \%) for x-vectors, i-vectors and CNN embeddings used in DNN-SAT with minimum length threshold.}
    \label{fig:wer_min_len}
\end{figure}

We also show the WER performance in the DNN-SAT setup for all three types of embeddings for different lengths of the utterances used in decoding (Fig.~\ref{fig:wer_min_len}). It is interesting to see that even though deep CNN embeddings are better speaker representations for short utterances (Fig.~\ref{fig:plda_xvector}), this is not reflected in their performance in DNN-SAT for short utterances. Regardless of the utterance length, i-vectors and deep CNN embeddings perform similarly for the task of input feature normalization. For x-vectors, even though they outperform the other embeddings in a speaker recognition task for longer utterances ($>$ 3.5 sec), they do not outperform the other embeddings when used for DNN-SAT, regardless of the utterance length. This result confirms that better speaker representations do not correlate with more effective normalization in DNN-SAT. 

\begin{table}[t]
    \centering
    \begin{tabular}{l|r}
        \hline
        \emph{Embedding type} & WER \\
        \hline
        SI baseline & 28.3 \\ 
        CMN & 27.0 \\ \hline
        online i-vectors & 25.9 \\
        utterance i-vectors & \textbf{25.8} \\
        CNN embeddings (100D PCA) & 26.2 \\
        CNN embeddings (100D LDA) & \textbf{25.8} \\
        CNN embeddings (20D LDA) & 26.1 \\ 
        x-vectors (pretrained, AMI backend) & 26.3 \\
        \hline
    \end{tabular}
    \caption{WERs for DNN-SAT with different embeddings. Models are trained with a control layer used to shift the input features.}
    \label{tab:different_embeddings}
\end{table}

\section{Conclusions}
\label{sec:conclusions}

In this paper we investigated  embeddings for efficient DNN-SAT. We evaluated the influence of the flexibility of the mapping applied to the embeddings, as well as
transforming input features compared to hidden representations. We show that transforming hidden layers is not more effective than learning shifts to the input features. With this simplified approach and an appropriate training strategy, the main part of the network is updated in the normalized feature space.
Although using a multi-layer control network to normalize all hidden representations gave similar performance, the simpler approach of linearly shifting the input features with single layer activations learned from the embeddings should be preferred. 

We also evaluated the effect of the embedding type on DNN-SAT ASR performance. We used different types of the embeddings for DNN-SAT (with the same strategy for generating control parameters), and for a speaker recognition task. This dual analysis of the embeddings provided insight into the characteristics of the embeddings desirable for DNN-SAT. We found frame-level variation of the embeddings did not bring WER improvements -- utterance-level summaries were the most beneficial for DNN-SAT. An utterance summary has the potential to capture more attributes than just one (e.g. speaker identity) if it is not explicitly designed to capture a specific attribute.  Adding speaker discrimination to the utterance summary was useful, but we found that the best speaker discriminability of the embeddings did not correlate with the best performance in DNN-SAT. In fact, a more important characteristic of the DNN-SAT embeddings is the ability to capture additional utterance attributes, rather than focusing solely on speaker differentiation. This can be achieved by i-vectors and deep CNN embeddings, but not by x-vectors in the current extraction framework. Adding channel or acoustic condition discriminability to x-vectors, perhaps using multi-task learning, could improve their performance for DNN-SAT. 

Analyzing the embeddings used in DNN-SAT with metrics other than WER can give insight into the reasons for differences in performance. This work is a step towards explaining DNN acoustic model adaptive training using auxiliary representations. Understanding why one embedding is superior to the other is valuable also for designing new embeddings. Therefore, we plan to investigate the same embeddings in more tasks (e.g. noise or dialect recognition) to relate their characteristics to the performance in DNN-SAT, and also plan to extract alternative embeddings for DNN-SAT using the knowledge acquired in this work.


\bibliographystyle{IEEEbib}
\bibliography{refs}

\end{document}